\pdfoutput=1

\documentclass[11pt]{article}

\usepackage{acl}

\usepackage{times}
\usepackage{latexsym}

\usepackage[T1]{fontenc}

\usepackage[utf8]{inputenc}

\usepackage{microtype}

\usepackage{graphicx}
\usepackage{hyperref}
\usepackage{url}
\usepackage{multirow}
\usepackage{makecell}
\usepackage{subfigure}
\usepackage{color}
\usepackage{xspace}
\usepackage{amsmath}
\usepackage{adjustbox}
\usepackage{booktabs}
\usepackage{soul}
\usepackage{float}
\graphicspath{ {./imgs/} }

\usepackage{color}

\title{Learning Representations for Hyper-Relational Knowledge Graphs}

\author{\quad Harry Shomer$^{1}$ \hspace{.3cm} Wei Jin$^{1}$ \hspace{.3cm} Juanhui Li$^{1}$ \hspace{.3cm} Yao Ma$^{2}$ \hspace{.3cm} Jiliang Tang $^{1}$\\
  $^1$ Department of Computer Science, Michigan State University, Michigan, USA\\
  $^2$ Department of Computer Science, New Jersey Institute of Technology, New Jersey, USA\\
  {\tt \{shomerha, jinwei2, lijuanh1, tangjili\}@msu.edu} \\ {\tt yao.ma@njit.edu}
  }

\begin{document}

\maketitle

\begin{abstract}
    Knowledge graphs (KGs) have gained prominence for their ability to learn representations for uni-relational facts. Recently, research has focused on modeling hyper-relational facts, which move beyond the restriction of uni-relational facts and allow us to represent more complex and real-world information. However, existing approaches for learning representations on hyper-relational KGs majorly focus on enhancing the communication from qualifiers to base triples while overlooking the flow of information from base triple to qualifiers. This can lead to suboptimal qualifier representations, especially when a large amount of qualifiers are presented.  It motivates us to design a  framework that utilizes multiple aggregators to learn representations for hyper-relational facts: one from the perspective of the base triple and the other one from the perspective of the qualifiers. Experiments demonstrate the effectiveness of our framework for hyper-relational knowledge graph completion across multiple datasets. Furthermore, we conduct an ablation study that validates the importance of the various components in our framework. The code to reproduce our results can be found at \url{https://github.com/HarryShomer/QUAD}.
\end{abstract}

\section{Introduction}

\begin{figure}[t]
    \centering
    \includegraphics[width=0.8\columnwidth]{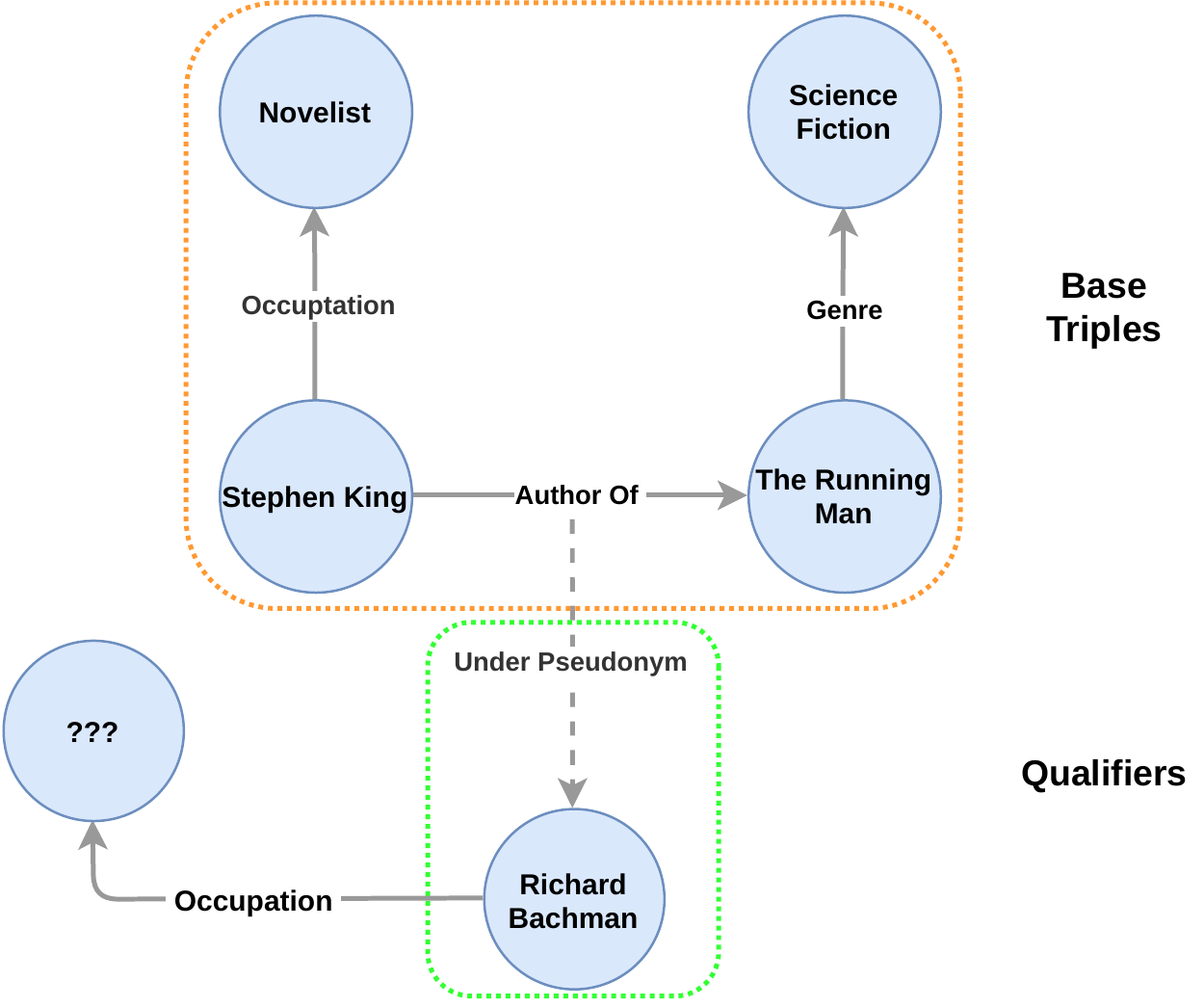}
    \caption{An example of a hyper-relational KG. The blue circles represent entities and the arrows represent relations. The dashed arrows represent qualifier relations while the solid arrows are base relations. The ??? entity represents the potential occupation of Richard Bachman (Novelist).
    }
    \label{fig:example}
\end{figure}

Knowledge graphs (KGs) are a collection of facts represented in a structured and graphical format. Facts are represented as a triple $(h, r, t)$ that connects two entities $h$ and $t$ (i.e. nodes) with a relation $r$. KGs have become very popular recently with applications in language representation \cite{liu2020k}, question answering \cite{huang2019knowledge}, and recommendation \cite{wang2019kgat}. 

In traditional triple-based KGs, facts are represented as binary relations between entities, which often fall short in representing the complex nature of the data. To address this shortcoming, hyper-relational KGs are introduced by moving from representing uni-relational facts to representing facts with N-ary relations. In hyper-relational KGs, triples are often associated with relation-entity pairs, which are known as \textit{qualifiers}. Qualifiers help qualify a given fact by providing more supporting information and are defined as an (entity, relation) pair that belongs to a triple. An example is shown in Figure \ref{fig:example}. In Figure \ref{fig:example} the triple \textit{(Stephen King, Author Of, The Running Man)} contains a single qualifier pair \textit{(Under Pseudonym, Richard Bachman)}. This qualifier pair helps provide more context to the base triple by telling us that Stephen King published the novel under the pseudonym Richard Bachman. Research on hyper-relational KGs~\cite{guan_nalp,galkin2020message,yu_hytransformer,wang_gran} have focused on learning representations for such hyper-relational graphs and examining how the addition of qualifier pairs can help boost the performance of knowledge graph completion. 

However, the majority of existing approaches only consider the impact from the qualifiers on base triples while overlooking the flow of information from the base triples to the qualifiers. This can lead to suboptimal performance especially when a large amount of qualifiers are presented. To illustrate its importance, we use Figure \ref{fig:example} as an example. In (\textit{Richard Bachman}, \textit{Occupation}, \textit{Novelist}), suppose the occupation of  \textit{Richard Bachman} is missing and we are trying to predict (\textit{Richard Bachman}, \textit{Occupation}, ?). 
If there is no information spreading from the base triple to the qualifier entity \textit{Richard Bachman}, predicting the missing link would be very hard as no knowledge of Stephen King is transferred to the qualifier entity. Prior work such as NaLP \cite{guan_nalp} and HINGE \cite{rosso_hinge} both struggle to achieve this transfer of information due to the simplicity of their convolutional-based frameworks. StarE \cite{galkin2020message} ignores such flow of information from the base triples to the qualifiers. Although transformer-based frameworks~\cite{yu_hytransformer,wang_gran} model the mutual influence between base triples and qualifiers via self attention, they inevitably ignore the structured nature of KGs.

Therefore, in this work, we aim to investigate the novel problem of learning representations for hyper-relational KGs by encouraging the mutual influence between base triples and qualifiers. Essentially, we are faced with the following challenge: how to  properly enhance the influence from base triples to qualifiers while maintaining effective impact from qualifiers to base triples.
To address it, we propose a novel framework - \textbf{QU}alifier Aggregated Hyper-Relation\textbf{A}l Knowle\textbf{D}ge Graphs (QUAD), which encourages influence in both directions.
Specifically, QUAD utilizes two aggregators from different perspectives - a base aggregator and a qualifier aggregator. The base aggregator aggregates  information for the base entities from the qualifiers while the qualifier aggregator performs the aggregation from the qualifier perspective. Inside the qualifier aggregator, we further propose the concept of a ``qualifier triple" that allows us to easily aggregate information from a base triple through the qualifier relation. 
Following both aggregations, the entity and relation representations are passed to the decoder to perform knowledge base completion.  
We show that our framework can achieve strong performance on multiple benchmark datasets. Our contributions can be summarized as follows:
\begin{itemize}
    \item We propose a novel architecture for hyper-relational knowledge graphs by introducing a graph encoder that aggregates information from the perspective of the qualifier entities.  
    \item We further show that several representative hyper-relational KG methods can be unified as the special cases of QUAD.
    \item  Extensive experiments demonstrate the effectiveness of our framework against numerous baselines on multiple hyper-relational knowledge graphs.
\end{itemize}


\begin{figure*}[t]
    \centering
    \includegraphics[width=6in]{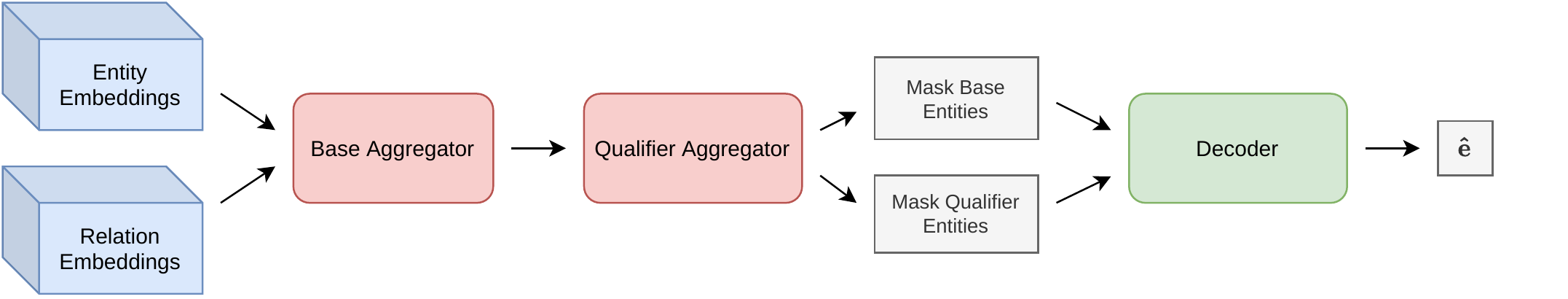}
    \caption{An overview of QUAD. It takes the set of entity and relation embeddings and encodes them using two aggregators. We then mask both the base and qualifier separately and pass the the statement to the decoder to predict the most likely entity $\hat{e}$ for the mask.}
    \label{fig:framework}
\end{figure*}

\section{Related Work}

\subsection{Knowledge Graph Embedding}

Knowledge graph embeddings (KGE) use embeddings to represent the latent features of entities and relations in a KG. Many KGE techniques also employ a score function that produce a score for a given triple $(h, r, t)$ measuring the plausibility of the triple being true \cite{ji2021survey}. Many different frameworks have been proposed in the literature. \citet{bordes2013translating} proposed using a translational scoring function to model triples. \citet{yang2014embedding} use a bilinear scoring function to score triples utilizing a diagonal matrix to model relations. RotatE \cite{sun2019rotate} scores a triple as a translation in a complex space and ConvE \cite{dettmers2018convolutional} does so using a convolutional neural network. Other works have focused on modeling knowledge graphs using graph neural networks. RGCN~\cite{schlichtkrull2018modeling} extends GCN \cite{kipf2016semi} by using a relation specific weight matrix when aggregating an entity's neighbors. However, RGCN suffers from over-parameterization when there are many relations. To alleviate this concern, CompGCN~\cite{vashishth2019composition} proposes to use direction specific weight matrices when aggregating instead of relational weight matrices. 

\subsection{Hyper-Relational KGs}
Several methods have been proposed that model hyper-relational facts as N-ary facts. \citet{wen2016_transh} propose m-Transh, a method that builds on TransH \cite{wang2014knowledge} by transforming each hyper-relational fact using a star-to-clique
conversion. RAE \cite{zhang2018_rae} builds upon m-TransH \cite{wen2016_transh} by further considering the relatedness between two entities. NaLP \cite{guan_nalp} uses a convolutional-based framework to compute a relatedness vector for a triple and its qualifier pairs that can be used for prediction. 

More recently, there has been some work that model hyper-relational KGs from a purely hyper-relational viewpoint.  \citet{rosso_hinge} propose doing so using a convolutional framework. For a given hyper-relational fact, the base triple is convolved by itself and with each specific qualifier pair. The resulting feature vectors are then combined and used for prediction. StarE \cite{galkin2020message} extends CompGCN \cite{vashishth2019composition} by encoding the qualifiers for a specific triple and combining it with the base relation of the triple. Using StarE \cite{galkin2020message} as their foundation, \citet{yu_hytransformer} replace the GNN aggregation with layer normalization layers \cite{ba2016layer} to improve performance. Additionally they mask the qualifier entities when training as a form of self-supervised learning (SSL). Lastly,  \citet{wang_gran} propose GRAN, a transformer based architecture that employs edge-specific attention biases and masks all entities and relations in the sequence. A missing element in previous work is the lack of a clear flow of information from the base triples to the qualifiers. To address this, we propose a graph encoder to aggregate information from the perspective of the qualifiers, thereby creating better qualifier encodings. 

\section{Preliminaries}

\subsection{Knowledge Graphs}

Let $\mathcal{G}=\{\mathcal{V}, \mathcal{R}, \mathcal{E} \}$ be a knowledge graph with nodes (i.e. entities) $\mathcal{V}$, edges $\mathcal{E}$, and relations $\mathcal{R}$. For $e \in \mathcal{E}$, it represents a directed edge where two entities $u\in\mathcal{V}$ and $v\in\mathcal{V}$ are connected by a relation $r\in\mathcal{R}$. We also denote the edge as a triple $(v, r, u)$. Further, we use $\mathcal{N}_v$ to denote the neighboring entities and relations of a node $v$.

CompGCN \cite{vashishth2019composition}, a popular method for modeling knowledge graphs,  utilizes a direction specific weight matrix $W_{\lambda(r)}$ and a function $\phi$ that combines the neighboring entity $\mathbf{h}_u$ and relation $\mathbf{h}_r$ of a given edge
\begin{equation}
    \mathbf{h}_v^{(k+1)} = f \left( \sum_{(u, r) \in \mathcal{N}_v} W_{\lambda(r)}^k \: \phi(\mathbf{h}_u^k, \mathbf{h}_r^k) \right),
\end{equation}
where $f$ is a non-linear function (e.g. ReLU) and $\lambda(r)$ represents the direction of relation that can be one of: a standard, inverse, or self-loop relation. Several function are proposed for modeling the interaction of the embedding in $\phi$ including subtraction, multiplication, and the cross correlation \cite{vashishth2019composition}. The relation embedding is updated through a transformation by a weight matrix $\mathbf{W}_{rel}$,
\begin{equation} \label{comp_rel_update}
    \mathbf{h}_r^{k+1} = \mathbf{W}_{rel} \: \mathbf{h}_r^{k} . 
\end{equation}


\subsection{Hyper-Relational Knowledge Graphs}

A hyper-relational knowledge graph can be seen as an extension of a standard knowledge graph where there is a set of qualifier pairs $\{(qv_i , qr_i)\}$, where $qv \in \mathcal{V}$ and $qr \in \mathcal{R}$, associated with each triple $(v, r, u)$.  For simplicity we use $q$ to represent the set of neighboring qualifier pairs  $(qv_i , qr_i) \in Q_{(v, r, u)}$ associated with a triple $(v, r, u)$. We can therefore represent a hyper-relational fact as $(v, r, u, q)$. We refer to hyper-relational facts as statements. Furthermore, we can represent the neighborhood for a qualifier entity $qv$ as $(v, r, u, qr) \in \mathcal{N}_{qv}$.

A representative example for modeling hyper-relational KGs is StarE~\cite{galkin2020message}. It proposes a formulation based on CompGCN to incorporates an embedding $\mathbf{h}_q$ that is an encoded representation of the qualifier pairs for the base triple $(v, r, u)$,
\begin{align} \label{stare_eq}
    &\mathbf{h}_v = f \left( \sum_{(u, r) \in \mathcal{N}_v} W_{\lambda(r)} \: \phi \left(\mathbf{h}_u, \gamma(\mathbf{h}_r, \mathbf{h}_q) \right) \right).
\end{align}
The embedding $\mathbf{h}_q$ is combined with the relation embedding $\mathbf{h}_r$ through a function $\gamma$ that performs a weighted sum. \citet{yu_hytransformer} show that replacing the graph aggregation with layer normalization can achieve comparable if not better performance than other frameworks. Furthermore they show  that masking the qualifier entities during training can raise the test performance on knowledge graph completion.

\subsection{Knowledge Graph Completion}

Knowledge graph completion (i.e. link prediction) masks one of the two entities belonging to the triple and attempts to predict the correct entity. For example, given $(v, r, u)$ we would try to predict the correct entity for both $(v, r, ?)$ and $(?, r, u)$. This is defined similarly for hyper-relational KGs where we are also provided with the qualifier information for the triple. Therefore, given $(v, r, u, q)$ we would try to predict the correct entity for both $(v, r, ?, q)$ and $(?, r, u, q)$.

\section{The Proposed Framework}

In this section, we propose a new framework QUAD for learning representations of hyper-relational knowledge graphs. An overview of QUAD is shown in Figure \ref{fig:framework} that consists of two main components. The first component, detailed in Section \ref{encoder_section}, encodes the graph using two separate aggregations while the second component detailed in Section \ref{decoder_section} decodes a given hyper-relational fact for a particular downstream task. In detail, we first pass the initial entity and relation embeddings to the encoder, which is comprised of two graph encoders that aggregate information from the base entities and qualifier entities, respectively. Once encoded, we create a separate sample for each statement masking each of the base and qualifier entities. Each sample then gets passed to the decoder to predict the most likely entity $\hat{e}$ for the mask.

\subsection{The Encoder} \label{encoder_section}
In this subsection, we define the encoder used in QUAD. The encoder is composed of two neighborhood aggregations: one that aggregates information for the base entities and one that does so for the qualifier entities. We refer to these two aggregators as the base aggregator and qualifier aggregator, respectively. The initial entity $E$  and relation $R$ embeddings are first passed to the base aggregator and then the qualifier aggregator. The encoded 
entity embedding $\hat{E}$ and relation embedding $\hat{R}$ can be expressed as follows:
\begin{align}
    E', R'   &= \text{Base-Agg} (E, R, \mathcal{G}) 
    , \\
    \hat{E}, \hat{R}   &= \text{Qual-Agg} (E', R', \mathcal{G}) \label{qual_agg_eq}, 
\end{align}
where $\text{Base-Agg}(\cdot)$ and $\text{Qual-Agg}(\cdot)$ are the aggregation functions in base aggregator and qualifier aggregator, respectively.
In the following, we introduce the details of both aggregators.

\subsubsection{The Base Aggregator} \label{base_encoder_section}

The base aggregator aims to  aggregate information for the base entities. Concretely, it takes $E$ and $R$  as input and aggregates the neighborhood information for a given base entity $v$ using a function $\psi_v$:
\begin{equation}
    \mathbf{h}_v = \text{Aggregate}(\psi_v(\mathbf{h}_u, \mathbf{h}_r, \mathbf{h}_q), \forall (u, r, q) \in \mathcal{N}_v),
\label{eq:base_agg}
\end{equation}
where $\mathbf{h}_q$ is the encoded representation for all the qualifier pairs belonging to the base triple $(v, r, u)$. We utilize CompGCN \cite{vashishth2019composition} as the Aggregate($\cdot$) function.
We can then rewrite Eq.~\eqref{eq:base_agg} as follows, 
\begin{equation} \label{base_agg_eq}
    \mathbf{h}_{v}  = f\left(\sum_{(u, r) \in \mathcal{N}(v)} \mathbf{W}_{\lambda(r)} \psi_v \left(\mathbf{h}_u, \mathbf{h}_r, \mathbf{h}_{q} \right) \right),
\end{equation}
where the encoded qualifier representation $\mathbf{h}_q$ is defined as in StarE \cite{galkin2020message} as:
\begin{equation}
    \mathbf{h}_{q} = \mathbf{W}_q \sum_{\left(qr, qv \right) \in Q_{(v, r, u)}} \phi \left(\mathbf{h}_{qr}, \mathbf{h}_{qv} \right),
\end{equation}
with $\mathbf{W}_q$ as the projection matrix, $\mathbf{h}_{qr}$ as the qualifier relation embedding, and $\mathbf{h}_{qv}$ as the qualifier entity embedding. Note that the relation $\mathbf{h}_r$ is updated through a linear transformation as shown in Eq. \eqref{comp_rel_update}.
StarE implements $\psi_v$ in Eq. \eqref{base_agg_eq} by combining the encoded qualifier information $\mathbf{h}_q$ with the triple's relation, i.e., $\psi_v(\mathbf{h}_u, \mathbf{h}_r, \mathbf{h}_{q})=\phi \left(\mathbf{h}_u, \gamma(\mathbf{h}_r, \mathbf{h}_q) \right)$ as in Eq.~\eqref{stare_eq}.
However, it is restricted by the assumption that qualifier information should be incorporated into the base relation embedding. Instead, we remove this restriction and combine the qualifier information with the output of the composition function $\phi$ as follows:
\begin{equation} \label{base_combine}
    \psi_v (\mathbf{h}_u, \mathbf{h}_r, \mathbf{h}_{q}) = \alpha \odot \phi(\mathbf{h}_{u}, \mathbf{h}_r) + (1 - \alpha) \odot \mathbf{h}_q,
\end{equation}
where $\alpha \in [0, 1]$ is a hyperparameter that balances the contribution of the base triple information and the encoded qualifiers.  In this way, the qualifier information encoded in $\mathbf{h}_q$ can directly interact with both the base entity and relation. 

\subsubsection{The Qualifier Aggregator} \label{qual_encoder_section}


The previously introduced base aggregator only considers the aggregation of information from the qualifiers to the base triple but not vice versa. Encoding base triple information in the qualifiers is advantageous as it can help learn better representations for the qualifiers. 
For example, in Figure \ref{fig:example} the base aggregation doesn't consider the flow of information from the triple \textit{(Stephen King, Author Of, The Running Man)} to the the qualifier entity \textit{Richard Bachman} resulting in limited information about the author being encoded in the qualifier entity. This makes it difficult to infer new facts where \textit{Richard Bachman} is a base entity. To encode more information for a qualifier entity, we can aggregate information from its neighbors, i.e., the base triples it belongs to and the qualifier relation connecting them to those triples. Using our previous example, the qualifier entity \textit{Richard Bachman} would aggregate information from the base triple \textit{(Stephen King, Author Of, The Running Man)} and the qualifier relation \textit{Under Pseudonym}. 

Hence, the Qual-Agg($\cdot$) function is designed to aggregate the neighborhood information for the qualifier entity $qv$ as follows:
\begin{align} \label{qual-agg}
    \mathbf{h}_{qv} = \: \text{Aggregate}(& \psi_q(\mathbf{h}_v, \mathbf{h}_r, \mathbf{h}_u,    \mathbf{h}_{qr}), \nonumber \\
    & \forall (v, r, u, qr) \in \mathcal{N}_{qv}),
\end{align}
where $\mathbf{h}_{qr}$ is the qualifier relation embedding and the function $\psi_q$ is used to combine the neighboring embeddings. For $\psi_q$, we hope that the base triple embeddings $(\mathbf{h}_v, \mathbf{h}_r, \mathbf{h}_u)$ should be considered as a whole. Since the qualifiers serve as additional context for explaining the whole triple, we should aggregate information from the whole triple instead of treating $\mathbf{h}_v, \mathbf{h}_r, \mathbf{h}_u, \mathbf{h}_{qr}$ individually. To achieve this goal, we first consider that a qualifier entity $qv$ is linked to some base triple $t = (v, r, u)$ by the qualifier relation $qr$. This can be seen as analogous to a standard triple where the base triple is the head, the qualifier relation is the relation, and the qualifier entity is the tail entity. For convenience, we can write this in a triple notation as $\left(t, qr, qv \right)$, which we refer to as a \textit{qualifier triple}. An example found in Figure \ref{fig:example} would be $t =$ \emph{(Stephen King, Author Of, The Running Man)}, $qr = \emph{Under Pseudonym}$ and $qv = \emph{Richard Bachman}$. Using these ideas, we can view $\psi_q$ as a function of the base triple $t$ and the qualifier relation $qr$:
\begin{equation}
    \psi_q(\mathbf{h}_v, \mathbf{h}_r, \mathbf{h}_u, \mathbf{h}_{qr}) = \phi(\mathbf{h}_{t}, \mathbf{h}_{qr} ),
\end{equation}
where the embedding $\mathbf{h}_t$ is the encoded representation of the base triple $t$ and $\phi$ is defined similarly to the composition function used in CompGCN \cite{vashishth2019composition}. The base triple representation $\mathbf{h}_t$ is formulated as the linear projection of the concatenated embeddings of the base triple:
\begin{equation}
    \mathbf{h}_t = \text{Linear} \left( \text{Concat} \left(\mathbf{h}_v , \mathbf{h}_r , \mathbf{h}_u \right) \right).
\end{equation}

\subsection{The Decoder} \label{decoder_section}
 Using the encoded representations of the entities and relations, i.e., $\hat{E}$ and $\hat{R}$, each hyper-relational fact is passed to a decoder to make the final prediction. To decode each fact we utilize a transformer \cite{vaswani2017attention} that employs an architecture similar to CoKE \cite{wang2019coke} extended to include qualifier information. For a given input sample $S$, we mask the entity token we are trying to predict. Eq. \eqref{decoder_mask} is an example where we mask the object entity:
\begin{equation} \label{decoder_mask}
    S = (\mathbf{h}_v, \: \mathbf{h}_r, \: [Mask], \: \mathbf{h}_{qr1}, \: \mathbf{h}_{qv1}, \: \cdots).
\end{equation}

After being passed through the transformer, we extract the masked embedding and pass it through a fully-connected layer and score it against all possible entities. This is then passed through a sigmoid function with the highest scoring entity being chosen as our prediction.

\subsubsection{Model Training} \label{training_section}
As the downstream task is to perform knowledge graph completion for entities in the base triples, we mask the head and tail entities  for the fact $(\mathbf{h}_v, \mathbf{h}_r, \mathbf{h}_u, q)$ where $q$ is the set of qualifier pairs associated with the triple. Then we try to predict the masked entities and minimize the loss $\mathcal{L}_{\text{base}}$ using binary cross-entropy loss.

To further enhance the learned representations of the embeddings and  exploit the qualifier information,  we include an auxiliary task that masks and attempts to predict the qualifier entities for each fact as proposed in Hy-Transformer \cite{yu_hytransformer}. We refer to this loss as $\mathcal{L}_{\text{qual}}$ and minimize it using the binary cross-entropy loss. Since the downstream task is solely to predict the missing base entities,  we introduce a hyperparameter $\beta \in [0, 1]$ that balances the contribution of the qualifier entity loss. The loss can now be written as thus:
\begin{equation}
    \mathcal{L} = \mathcal{L}_{\text{base}} + \beta \mathcal{L}_{\text{qual}}.
\end{equation}

\subsection{Parallel Architecture} \label{parallel_section}
We also consider a version of QUAD that combines the entity representations encoded by the two aggregation schemes in parallel instead of sequentially. Under this setting, both aggregate functions take the initial entity embedding matrix $\mathcal{E}$ as input. The encoded entity representations outputted by the two encoders are then combined via a weight matrix $\mathbf{W}_{p}$ and passed to the decoder. The relation embeddings are still passed sequentially as we found that this performed best. We believe that this version of our framework may be better at balancing the contribution of the base and qualifier aggregation for some datasets. We formulate the parallel encoding scheme as follows where $E_{base}$ and $R_{base}$ are the output of $\text{Base-Agg}(\cdot)$ and $E_{qual}$ the output of $\text{Qual-Agg}(\cdot)$:
\begin{align}
    &E_{base}, R_{base} = \text{Base-Agg} (E, R, \mathcal{G}), \\
    &E_{qual}, \hat{R}  = \text{Qual-Agg} (E, R_{base}, \: \mathcal{G}), \\
    &\hat{E}  = \mathbf{W}_{p} \; \text{Concat} \left( E_{base}, \: E_{qual} \right).
\end{align}

\subsection{Relationship to Other Frameworks}

\begin{table*}[t]
\small
	\centering
	\caption{Knowledge Graph Completion Results for Multiple Datasets}
	\label{tab:general_results}
	\adjustbox{max width=\linewidth}{
		\begin{tabular}{@{}lcccccccccccc@{}}
			\toprule
			   \multicolumn{1}{c}{\textbf{Method}} &
			   \multicolumn{3}{c}{\textbf{WD50K (13.6) }} & \multicolumn{3}{c}{\textbf{Wikipeople (2.6)}} & \multicolumn{3}{c}{\textbf{JF17K (45.9)}}  \\ 
			   \cmidrule(l){2-4} \cmidrule(l){5-7} \cmidrule(l){8-10} 
			   & MRR   & H@1   & \multicolumn{1}{c}{H@10} & MRR   & H@1   & \multicolumn{1}{c}{H@10} & MRR   & H@1   & \multicolumn{1}{c}{H@10}  
			   \\ \midrule

			 NaLP-Fix & 0.177 & 0.131 & 0.264   & 0.420 & 0.343 & 0.556   & 0.245 & 0.185  & 0.358 \\
			 HINGE    & 0.243 & 0.176 & 0.377   & 0.476 & 0.415  & 0.585     & 0.449 & 0.361 &  0.624 \\
			 StarE    & 0.349 & 0.271 & 0.496   & 0.491  & 0.398 & \textbf{0.648}    & 0.574 & 0.496  &  0.725 \\
			 Hy-Transformer  & \textbf{0.356} & \textbf{0.281} & \textbf{0.498}  & \textbf{0.501} & 0.426 & 0.634    & 0.582 & 0.501 & 0.742 \\
			 \hline
			 QUAD     & 0.348 & 0.270 & 0.497   & 0.466 & 0.365 & 0.624    & 0.582 & 0.502 & 0.740 \\
			 QUAD (Parallel) & 0.349 & 0.275 & 0.489    & 0.497 & \textbf{0.431} & 0.617    & \textbf{0.596} & \textbf{0.519} & \textbf{0.751} \\ 
			\bottomrule
		\end{tabular}}
\end{table*}

\begin{table*}[t]
\small
	\centering
	\caption{Knowledge Graph Completion Results on WD50K Splits}
	\label{tab:WD50K_results}
	\adjustbox{max width=\linewidth}{
		\begin{tabular}{@{}lcccccccccccc@{}}
			\toprule
    		   \multicolumn{1}{c}{\textbf{Method}} &
			   \multicolumn{3}{c}{\textbf{WD50K (33)}} & \multicolumn{3}{c}{\textbf{WD50K (66)}} & \multicolumn{3}{c}{\textbf{WD50K (100)}} \\ \cmidrule(l){2-4} \cmidrule(l){5-7} \cmidrule(l){8-10} 
			     & MRR   & H@1   & \multicolumn{1}{c}{H@10} & MRR   & H@1   & \multicolumn{1}{c}{H@10} & MRR   & H@1   & H@10  \\ \midrule

			 NaLP-Fix & 0.204 & 0.164 & 0.277  & 0.334 & 0.284 & 0.423  & 0.458 & 0.398 & 0.563 \\
			 HINGE    & 0.253 & 0.190 & 0.372  & 0.378 & 0.307 & 0.512  & 0.492 & 0.417 & 0.636 \\
			 StarE    & 0.331 & 0.268 & 0.451  & 0.481 & 0.420 & 0.594  & 0.654 & 0.588 & 0.777 \\
			 Hy-Transformer & 0.343 & - & - & \textbf{0.515} & - & - & 0.699 & 0.637	& 0.812 \\
			 \hline
			 QUAD     & \textbf{0.349} &  0.286 & \textbf{0.470}      & \textbf{0.515} & \textbf{0.456} & \textbf{0.623}     & \textbf{0.703} & \textbf{0.638} & \textbf{0.820} \\
			 QUAD (Parallel) & 0.346 & \textbf{0.287} & 0.459    & 0.510 & 0.454 & 0.615    & 0.693 & 0.628 & 0.812 \\ 		 
			\bottomrule
		\end{tabular}}
\end{table*}

In this subsection we provide a unified view on several popular hyper-relational KG methods by showing that they are subsets of our framework.

\subsubsection{StarE}

StarE \cite{galkin2020message} uses a GNN to encode the KG and a transformer to perform knowledge graph completion. The GNN encoder is detailed in Eq. \eqref{stare_eq}. We can show that StarE is a special case of QUAD where:
\begin{itemize}
    \item Information is only aggregated from the perspective of the base entities. $\text{Qual-Agg}(\cdot)$ is therefore equal to the identity function.
    \item The function $\psi_v$ in the base aggregator is defined as $\phi(\mathbf{h}_u, \alpha \odot \mathbf{h}_r + (1 - \alpha) \odot \mathbf{h}_q)$.
    \item StarE only masks the base entities so $\beta=0$.
\end{itemize}

\subsubsection{Hy-Transformer}

Hy-Transformer \cite{yu_hytransformer} modifies StarE by replacing the GNN encoder with layer normalization layers. They also introduce an auxiliary task that masks the qualifier entities during training. We demonstrate that Hy-Transformer is a special case of QUAD where:
\begin{itemize}
    \item Both aggregate functions are equal to layer normalization layers for the entity and relation embeddings such that: $\hat{\mathcal{E}} = \text{Layer-Norm}(\mathcal{E})$ and $\hat{\mathcal{R}} = \text{Layer-Norm}(\mathcal{R})$. 
    \item The loss balancing hyperparameter $\beta = 1$. 
\end{itemize}

\subsubsection{GRAN}
GRAN \cite{wang_gran} uses a transformer with edge-specific attention biases. We demonstrate that GRAN is a special case of QUAD where:
\begin{itemize}
    \item Both aggregate functions are  equal to the identity function.
    \item The loss balancing hyperparameter $\beta=1$.
    \item The transformer (decoder) considers edge-specific biases. The transformer used in QUAD is equivalent to the GRAN-complete model where the biases for the key and value matrices are set to zero $(e_{ij}^K, e_{ij}^V) = (0, 0)$.
\end{itemize}

\section{Experiment}

In this section, we conduct experiments to demonstrate the effectiveness of our proposed framework QUAD. We first introduce the experimental settings and then compare the results of QUAD against the baselines on numerous benchmark datasets. Next we perform an ablation study to determine the importance of each component in our framework. Lastly, we perform some additional experiments to assess the impact of the loss balancing term on the performance. 

\subsection{Experimental Settings}

\subsubsection{Datasets} \label{experiment_datasets}

We consider three datasets for our experiments including JF17K \cite{wen2016_transh}, Wikipeople \cite{guan_nalp}, and WD50K \cite{galkin2020message}.
An issue with WD50K and Wikipeople is that only a small percentage of triples contain qualifiers, being 13.6\% for WD50K and 2.6\% for Wikipeople. We therefore also measure the performance on the WD50K splits introduced by \citet{galkin2020message} that contain a higher percentage of triples with qualifiers. The three splits are WD50K (33), WD50K (66), and WD50K (100) with the number in parentheses representing the percentage of triples with qualifiers.

\subsubsection{Baselines}
  
We compare the results of our framework with other prominent hyper-relational baselines including 
NaLP-Fix \cite{rosso_hinge}, HINGE \cite{rosso_hinge}, StarE \cite{galkin2020message}, and Hy-Transformer \cite{yu_hytransformer}. Note that we do not include GRAN \cite{wang_gran} as one baseline since (1) similar to Hy-Transformer, it is also a transformer-based method; and (2) in addition to the auxiliary task, GRAN also masks the relations and we can incorporate such component to the proposed framework that we leave as one future work.  

\subsubsection{Evaluation Metrics}
To evaluate the performance on the test set we report the mean reciprocal rank (MRR) and the percentage of top 1 and 10 hits (H@1 and H@10) when performing knowledge graph completion. We utilize the filtered setting introduced by \citet{bordes2013translating}.

\subsection{Performance Comparison on Benchmarks}

\subsubsection{Performance on WD50K, Wikipeople and JF17K}

In this subsection, we evaluate QUAD on the benchmark datasets and compare its performance to the aforementioned baselines. We first evaluate performance on the WD50K, Wikipeople, and JF17K datasets. The method \textit{QUAD} is the original formulation of our framework while the method \textit{QUAD (Parallel)} is the alternative formulation presented in Section \ref{parallel_section}. The results are shown in Table \ref{tab:general_results}. For each dataset we include the percentage of triples with qualifiers in parentheses. 

Evaluating the results in Table \ref{tab:general_results} we observe that the performance of QUAD varies by dataset. For JF17K it achieves the best performance for all three metrics including a 2.4\% increase in MRR over the second best performing model. On Wikipeople its performance is similar Hy-Trasnformer while for WD50K it is slightly below state of the art. This is due to both WD50K and Wikipeople containing a low percentage of triples with qualifier pairs with 13.6\% and 2.6\%, respectively. On JF17K, which has a much higher percentage of qualifiers at 45.9\%, QUAD is able to outperform the baseline models. We therefore believe that datasets with a higher ratio of qualifiers is where QUAD shows its value. 

\subsubsection{Varying the ratio of qualifiers}

To test this hypothesis, we evaluate QUAD on the WD50K subsets introduced in Section \ref{experiment_datasets}. The percentage of triples with qualifier pairs for the three subsets is approximately 33\%, 66\%, and 100\%, respectively. The results are presented in Table \ref{tab:WD50K_results}. From the results, we see state-of-the-art performance across all the three datasets for each of the three evaluation metrics. Interestingly, as opposed to the results shown in Table \ref{tab:general_results} the original (non-parallel) version of QUAD performs the best. We believe that the strong performance on the datasets with a more substantive percentage of qualifiers shows the utility of our framework and that the relatively poorer performance on the Wikipeople and WD50K datasets result from them containing too few triples with qualifiers.

\subsection{Ablation Study}

In this subsection, we conduct an ablation study to determine the importance for each component in QUAD. We do so by considering three versions of QUAD: (1) One that doesn't mask the qualifier entities, (2) one without the qualifier aggregation component and (3) one without (1) and (2). Evaluating the results on the three ablated frameworks will help us ascertain both the individual and cumulative effect those two components have on the performance. Of most importance is the impact of the novel qualifier aggregation introduced in this paper. We report the results of this study on the WD50K (100) dataset under the original (non-parallel) setting. The results can be found in Table~\ref{tab:ablation_results}. 

\begin{table}[h]
\small
    \centering
    \caption{Ablation Study on WD50K (100)}
    \label{tab:ablation_results}
    \begin{tabular}{@{}lccccc@{}}
        \toprule
            \textbf{Method}  & 
            \textbf{MRR} & 
            \textbf{H@1} & 
            \textbf{H@10} \\
        \midrule
       
        w/o Qual Agg \& Mask  & 0.658 & 0.591 & 0.781  \\
        w/o Qual Mask         & 0.677 & 0.613 & 0.794  \\
        w/o Qual Agg          & 0.696 & 0.628 & \textbf{0.820}  \\
        QUAD                  & \textbf{0.703} & \textbf{0.638} & \textbf{0.820} \\
        \bottomrule
    \end{tabular}
\end{table}

Evaluating the results in Table \ref{tab:ablation_results} we see that the removal of either or both of the two components leads to a degradation in performance. Ablating the  qualifier aggregation, the main contribution of our framework, causes a 1\% reduction in MRR and a 1.6\% drop in H@1. Furthermore, removing just the qualifier masking results in a 3.7\% reduction in MRR and ablating both components leads to the largest drop with a 6.4\% drop in MRR. These results validate the importance of the qualifier aggregation as its removal leads to a degradation in performance.

\subsection{Effect of Loss Balancing Hyperparameter}

In this subsection, we study the effect that the loss balancing hyperparameter $\beta$ has on the performance. To do so we train and evaluate multiple versions of our framework with values of $\beta$ in the set $\{0, 0.25, 0.5, 0.75, 1\}$. We note that a value of $\beta=0$ is equivalent to no qualifier masking while a value of $\beta=1$ is equivalent to the loss used in Hy-Transformer \cite{yu_hytransformer}. We perform this study on the three WD50K splits and JF17K to determine the impact the loss balancing hyperparameter has on a variety of datasets. We utilize the parallel setting when training on JF17K and our original formulation on the WD50K splits. For simplicity we report only the MRR performance in Table \ref{tab:balancing_results}.

\begin{table}[h]
\small
    \centering
    \caption{Loss Hyperparameter Study (MRR)}
    \label{tab:balancing_results}
    
    \begin{tabular}{p{8mm} p{11mm} p{11mm} p{11mm} p{11mm}}
    
        \toprule
            $\mathbf{\beta}$  & 
            \textbf{WD50K (33)} & 
            \textbf{WD50K (66)} & 
            \textbf{WD50K (100)} & 
            \textbf{JF17K}  \\ 
        \midrule
       0    &     0.33  &    0.491 &     0.677 &     0.592  \\ 
	   0.25 &     0.346 &    0.507 &     0.690 &     \textbf{0.596}  \\ 
	   0.50 &     0.348 &    0.514 &     0.697 &     0.592  \\
	   0.75 &     0.348 &    0.514 &     0.701 &     0.589  \\
	   1    &     \textbf{0.349} &    \textbf{0.515} &     \textbf{0.703} &     0.583  \\ 
        \bottomrule
    \end{tabular}
\end{table}

The results in Table \ref{tab:balancing_results} shows multiple trends. For WD50K (100) there is a clear relationship with the value of $\beta$ and the performance. The higher the value of $\beta$ the better the performance. This is likely due to the fact that every triple in the dataset has at least one qualifier pair. For both WD50K (33) and (66) the MRR improves as we increase $\beta$ from 0 to 0.50 but we see little to no improvement when increasing the value from 0.50 to 1. Lastly, for JF17K the MRR is maximized when $\beta = 0.25$. Furthermore, for increasing values of $\beta$ beyond 0.25 we see a noticeable drop in performance resulting in a MRR 2.2\% lower at $\beta=1$ compared to $\beta=0.25$. This shows the importance of balancing the magnitude of the auxiliary loss, as depending on the dataset balancing them equally ($\beta = 1$) can potentially lead to a large degradation in performance.

\section{Conclusion}

In this paper we introduce our framework QUAD for learning representations for hyper-relational knowledge graphs. Our framework is motivated by better encoding qualifier information for a given hyper-relational fact. To this point we design a novel qualifier aggregation module to learn better encoded representations for the qualifiers. We also introduce a hyperparameter to balance the contribution of the auxiliary task in the loss. Experiments show that our framework performs well on several benchmark datasets as compared to competitive baselines. Further experiments validate the importance of the various components in our framework and the need to balance the auxiliary loss. For future work we plan on exploring how additional auxiliary information such as text description or literals help learn better representations of hyper-relational KGs.

\bibliography{anthology, references}
\bibliographystyle{acl_natbib}

\clearpage
\appendix

\section{Infrastructure}

All experiments were done on one 32G Tesla V100 GPU and implemented using Pytorch Geometric \cite{torch_geometric}.

\section{Datasets}

We conducted experiments on three datasets. These are JF17K \cite{wen2016_transh}, Wikipeople \cite{guan_nalp}, and WD50K \cite{galkin2020message}. Table  \ref{tab:datasets} details the statistics for each.

\begin{table}[h]
\centering
\footnotesize
    \caption{Datasets. We note the approximate percentage of triples with qualifiers in parentheses.}
    \begin{tabular}{@{}lcccc@{}}
        \toprule
            \textbf{Dataset}  & 
            \multicolumn{1}{l}{\textbf{Statements}} & 
            \multicolumn{1}{l}{\textbf{\#Entities}} & 
            \multicolumn{1}{l}{\textbf{\#Relations }} \\ 
        \midrule
        
        Wikipeople (2.6) & 369,866 & 34,839 & 375 \\
         \midrule \midrule
        JF17K (45.9)      & 100,947 & 28,645 & 501 \\
        \midrule \midrule
        WD50K (33) & 102,107 & 38,124 & 475   \\
        WD50K (66) & 49,167 & 27,347 & 494   \\
        WD50K (100) & 31,314 & 18,792 & 279   \\
        WD50K (13.6) & 236,507 & 47,156 & 532  \\
        \bottomrule
    \end{tabular}
\label{tab:datasets}
\end{table}

\section{Hyperparameters}

QUAD is trained for 500 epochs with a learning rate of 1e-4. Furthermore the embedding dimension is 200 and the number of layers for the Base-Agg encoder is 2. For the transformer we set the number of layers to 2, the number of heads to 4, dropout to 0.1, and the hidden dimension is tuned from \{512, 768\}. The batch size is tuned from \{128, 256\}, $\alpha$ is tuned from [0, 1] in steps of 0.1, beta is tuned from \{0.25, 0.5, 0.75, 1\}, the number of layers for the Qual-Agg from \{1, 2\}, the encoding dropout from \{0.1, 0.2, 0.3\}, the learning rate decay from \{None, 0.996, 0.9975, 0.999\}, and the label smoothing from \{0.1, 0.2, 0.4, 0.6, 0.8\}. The Adam optimizer is used in all the experiments \cite{kingma2014adam}. For the composition function $\phi$ we utilize the RotatE scoring function \cite{sun2019rotate}. Under the parallel setting we tune an additional dropout layer after we combine the representation of the two encoders from \{0.2, 0.3\}.

Table \ref{tab:general_hyperparams} holds the hyperparameters for the general datasets and Table \ref{tab:splits_hyperparams} for the three WD50K splits. The values are the same for the parallel and non-parallel versions of QUAD with the exception of the inclusion of the parallel dropout hyperparameter.

\begin{table}[h]
\small
    \centering
    \caption{Hyperparameter values for the general datasets.}
    \begin{tabular}{@{}lccc@{}}
        \toprule
            \textbf{Hyperparameter}  & 
            \multicolumn{1}{l}{\textbf{JF17K}} & 
            \multicolumn{1}{l}{\textbf{Wikipeople}} & 
            \multicolumn{1}{l}{\textbf{WD50K}} \\
        \midrule

        Batch Size       & 128 & 128 & 128 \\
        Alpha            & 0.8 & 0.8 & 0.7 \\
        Beta             & 0.25 & 0 & 0.5  \\
        Qual Agg Layers  & 2 & 2 & 1       \\
        Encoder Dropout  & 0.1 & 0.1 & 0.2    \\ 
        Transformer Dim  & 768 & 512 & 768 \\ 
        LR Decay         & 0.999 & 0.999 & 0.9975 \\
        Label Smoothing  & 0.6  & 0.2 & 0.2 \\
        Parallel Dropout & 0.2  & 0.2 & 0.2 \\
        
        \bottomrule
    \end{tabular}
    \label{tab:general_hyperparams}
\end{table}

\begin{table}[h]
\small
    \centering
    \caption{Hyperparameter values for the WD50K splits.}
    \begin{tabular}{@{}lccc@{}}
        \toprule
            \textbf{Hyperparameter}  & 
            \multicolumn{1}{c}{\textbf{(33)}} & 
            \multicolumn{1}{c}{\textbf{(66)}} &
            \multicolumn{1}{c}{\textbf{(100)}} \\
        \midrule

        Batch Size       & 128 & 128 & 256 \\
        Alpha            & 0.7 & 0.7 & 0.6 \\
        Beta             & 1 & 1 & 1 \\
        Qual Agg Layers  & 1 & 1 & 1 \\
        Encoder Dropout  & 0.2 & 0.2 & 0.2 \\ 
        Transformer Dim  & 768 & 768 & 768 \\ 
        LR Decay         & 0.9975 & 0.9975 & None \\
        Label Smoothing  & 0.2 & 0.2 & 0.1 \\
        Parallel Dropout & 0.3 & 0.3 & 0.3 \\
        
        \bottomrule
    \end{tabular}
    \label{tab:splits_hyperparams}
\end{table}

\end{document}